%% file: main.tex
\pdfoutput=1

\documentclass[11pt]{article}

\usepackage{acl}

\usepackage{times}
\usepackage{latexsym}

\usepackage[T1]{fontenc}

\usepackage[utf8]{inputenc}

\usepackage{microtype}

\usepackage{inconsolata}

\usepackage{color, colortbl}
\usepackage{adjustbox}
\usepackage{amsmath}
\definecolor{Gray}{gray}{0.9}

\usepackage[ruled,vlined]{algorithm2e}

\usepackage{xcolor}         
\usepackage{todonotes}
\usepackage{amssymb} 
\usepackage{bm}

\input{latex/custom_commands}

\def\vc{{\bm{c}}}

\def\vy{{\bm{y}}}
\def\vz{{\bm{z}}}

\def\mL{{\bm{L}}}

\def\mX{{\bm{X}}}

\def\mZ{{\bm{Z}}}

\author{Alexandre Audibert \\
Université Grenoble Alpes
   \\\And
Aurélien Gauffre \\
Université Grenoble Alpes
      \\\And
  Massih-Reza Amini \\
  Université Grenoble Alpes}

\title{Exploring Contrastive Learning for Long-Tailed Multi-Label Text Classification}

\date{September 2023}

\begin{document}

\maketitle
\input{latex/0Abstract}

\input{latex/1Intro}

\input{latex/2RelatedWork}

\input{latex/3OurWork}

\input{latex/4ExperimentationDetails}
\input{latex/5Results}

\input{latex/6Conclusion}
\input{latex/7Limitation}

\bibliography{custom}

\appendix
\input{latex/8Appendix}
\end{document}

%% file: latex/custom_commands.tex
\DeclareMathAlphabet{\mathsfit}{\encodingdefault}{\sfdefault}{m}{sl}
\SetMathAlphabet{\mathsfit}{bold}{\encodingdefault}{\sfdefault}{bx}{n}



%% file: latex/0Abstract.tex
\begin{abstract}

Learning an effective representation in multi-label text classification (MLTC) is a significant challenge in NLP. This challenge arises from the inherent complexity of the task, which is shaped by two key factors: the intricate connections between labels and the widespread long-tailed distribution of the data. To overcome this issue, one potential approach involves integrating supervised contrastive learning with classical supervised loss functions. Although contrastive learning has shown remarkable performance in multi-class classification, its impact in the multi-label framework has not been thoroughly investigated. In this paper, we conduct an in-depth study of supervised contrastive learning and its influence on representation in MLTC context. We emphasize the importance of considering long-tailed data distributions to build a robust representation space, which effectively addresses two critical challenges associated with contrastive learning that we identify: the "lack of positives" and the "attraction-repulsion imbalance". Building on this insight, we introduce a novel contrastive loss function for MLTC. It attains Micro-F1 scores that either match or surpass those obtained with other frequently employed loss functions, and demonstrates a significant improvement in Macro-F1 scores across three multi-label datasets.


\end{abstract}
%



%% file: latex/1Intro.tex
\section{Introduction}
In recent years, multi-label text classification has gained significant popularity in the field of Natural Language Processing (NLP). Defined as the process of assigning one or more labels to a document, MLTC plays a crucial role in numerous real-world applications such as document classification, sentiment analysis, and news article categorization.

Despite its similarity to multi-class mono-label text classification, MLTC presents two fundamental challenges: handling multiple labels per document and addressing datasets that tend to be long-tailed. These challenges highlight the inherent imbalance in real-world applications, where some labels are more present than others, making it hard to learn a robust semantic representation of documents. 

Numerous approaches have emerged to address this issue, such as incorporating label interactions in model construction and devising tailored loss functions. Some studies advocate expanding the representation space by incorporating statistical correlations through graph neural networks in the projection head \cite{gcn_springer, emograph}. Meanwhile, other approaches recommend either modifying the conventional Binary Cross-Entropy (BCE) by assigning higher weights to certain samples and labels or introducing an auxiliary loss function for regularization \cite{enhancing_multi-label}. Concurrently, recent approaches based on supervised contrastive learning employed as an auxiliary loss managed to enhance semantic representation in multi-class classification \cite{parametric, nlpclclass}. 

While contrastive learning represents an interesting tool, its application in MLTC remains challenging due to several critical factors. Firstly, defining a positive pair of documents is difficult due to the interaction between labels. Indeed, documents can share some but not all labels, and it can be hard to clearly evaluate the degree of similarity required for a pair of documents to be considered positive. Secondly, the selection of effective data augmentation techniques necessary in contrastive learning proves to be a non-trivial task. Unlike images, where various geometric transformations are readily applicable, the discrete nature of text limits the creation of relevant augmentations. Finally, the data distribution in MLTC often shows an unbalanced or long-tailed pattern, with certain labels being noticeably more common than others. This might degrade the quality of the representation \cite{dissecting,zhu2022balanced}. Previous research in MLTC has utilized a hybrid loss, combining supervised contrastive learning with classical BCE, without exploring the effects and properties of contrastive learning on the representation space. Additionally, the inherent long-tailed distribution in the data remains unaddressed, leading to two significant challenges that we term as ``lack of positive'' and ``attraction-repulsion imbalance''. The ``lack of positive'' issue arises when instances lack positive pairs in contrastive learning, and the ``attraction-repulsion imbalance'' is characterized by the dominance of attraction and repulsion terms for the labels in the head of the distribution.

In this paper, we address these challenges head-on and present a novel multi-label supervised contrastive approach, referred to as ABALONE, introducing the following key contributions:
\begin{itemize}
\item We conduct a comprehensive examination of the influence of contrastive learning on the representation space, specifically in the absence of BCE and data augmentation. 
\item We put forth a substantial ablation study, illustrating the crucial role of considering the long-tailed distribution of data in resolving challenges such as the ``Attraction-repulsion imBAlance'' and ``Lack of pOsitive iNstancEs''. 
\item We introduce a novel contrastive loss function for MLTC that attains Micro-F1 scores on par with or superior to existing loss functions, along with a marked enhancement in Macro-F1 scores.
\item Finally, we examine the quality of the representation space and the transferability of the features learned through supervised contrastive learning.
\end{itemize}

The structure of the paper is as follows: in Section \ref{sec:RW}, we provide an overview of related work. Section \ref{sec:OW} introduces the notations used throughout the paper and outlines our approach. In Section \ref{sec:Exps}, we present our experimental setup, while Section \ref{sec:Results} provides results obtained from three datasets. Finally, Section \ref{sec:Conclusion} presents our conclusions.

%% file: latex/2RelatedWork.tex
\section{Related Work}
\label{sec:RW}

In this section, we delve into an exploration of related work on supervised contrastive learning, multi-label text classification,  and the application of supervised contrastive learning to MLTC.

\subsection{Supervised Contrastive Learning}
The idea of supervised contrastive learning has emerged in the domain of vision with the work of \citet{SupCon} called \textit{SupCon}. This study demonstrates how the application of a supervised contrastive loss may yield results in multi-class classification that are comparable, and in some cases even better, to the traditional approaches. The fundamental principle of contrastive learning involves enhancing the representation space by bringing an anchor and a positive sample closer in the embedding space, while simultaneously pushing negative samples away from the anchor. In supervised contrastive learning, a positive sample is characterized as an instance that shares identical class with the anchor. In \citet{dissecting}, a comparison was made between the classical cross-entropy loss function and the \textit{SupCon} loss. From this study, it appeared that both loss functions converge to the same representation under balanced settings and mild assumptions on the encoder. However, it was observed that the optimization behavior of \textit{SupCon} enables better generalization compared to the cross-entropy loss.

In situations where there is a long-tailed distribution, it has been found that the representation learned via the contrastive loss might not be effective. One way to improve the representation space is by using class prototypes \cite{zhu2022balanced, parametric, dissecting}. Although these methods have shown promising results, they primarily tackle challenges in multi-class classification problems.

\subsection{Multi-label Classification}

Learning MLTC using the binary cross-entropy loss function, while straightforward, continues to be a prevalent approach in the literature. A widely adopted and simple improvement to reduce imbalance in this setting is the use of focal loss \cite{focal}. This approach prioritizes difficult examples by modifying the loss contribution of each sample, diminishing the loss for well-classified examples, and accentuating the importance of misclassified or hard-to-classify instances. An alternative strategy involved employing the asymmetric loss function \cite{asymmetric}, which tackles the imbalance between the positive and negative examples during training. This is achieved by assigning different penalty levels to false positive and false negative predictions. This approach enhances the model's sensitivity to the class of interest, leading to improved performance, especially in datasets with imbalanced distributions.
 
Other works combine an auxiliary loss function with BCE, as in multi-task learning, where an additional loss function serves as regularization. For instance, \citet{enhancing_multi-label} suggest incorporating an auxiliary loss function that specifically addresses whether two labels co-occur in the same document. Similarly, \citet{spanemo} propose a label-correlation-aware loss function designed to maximize the separation between positive and negative labels inside an instance.

Rather than manipulating the loss function, alternative studies suggest adjusting the model architecture. A usual approach involves integrating statistical correlations between labels using Graph Neural Network \cite{emograph, dual_mltc-gcn, gcn_springer}. Additionally, a promising avenue of research looks into adding label parameters to the model, which would enable the learning of a unique representation for every label as opposed to a single global representation \cite{decoder-encoder, spanemo, lsan}.

\subsection{Supervised Contrastive Learning for Multi-label Classification}
The use of supervised contrastive learning in multi-label classification has recently gained interest within the research community. All the existing studies investigate the effects of supervised contrastive learning by making some kind of prior assumption about label interactions in the learned representation space. 

\citet{Mulcon} suggest to use supervised contrastive learning for image classification based on the assumption that labels are situated in distinct areas of an image. Their contrastive loss is utilized alongside the BCE loss function and serves as a type of regularization more details can be found in Appendix \ref{appendix:mulocn}.

\citet{effectivemltcco} propose five different supervised contrastive loss functions that are used jointly with BCE to improve semantic representation of classes. In addition, \citet{contrastive_MLTC} suggest using a KNN algorithm during inference in order to improve performance. Some studies use supervised contrastive learning with a predefined hierarchy of labels \cite{hierarchicalContrastiveLearning, hiermltcl}. 

While contrastive loss functions in mono-label multi-class scenarios push apart representations of instances from different classes, directly applying this approach to the multi-label case may yield suboptimal representations, particularly for examples associated with multiple labels.  This can lead to a deterioration in results, particularly in long-tail scenarios.

In contrast to other methods, our approach does not rely on any prior assumptions about label interactions. We address the long-tail distribution challenge in MLTC by proposing several key changes in the supervised contrastive learning loss.

%% file: latex/3OurWork.tex
\section{ABALONE}
\label{sec:OW}
We begin by introducing the notations and then present our approach. In the following, $B$ is defined as the set of indices of examples in a batch, and $L$ represents the number of labels. The representation of the $i^{th}$ document in a batch is denoted as $\vz_{i}$. The associated label vector for example $i$ is $\vy_i \in \{0, 1\}^L$, with $y_i^j$ representing its $j^{th}$ element. 
Furthermore, we denote by $I_B=\{\vz_i\mid i\in B\}$ the set of document embeddings in the batch $B$.
\begin{figure*}[t]
 \centering
\includegraphics[scale=.62]{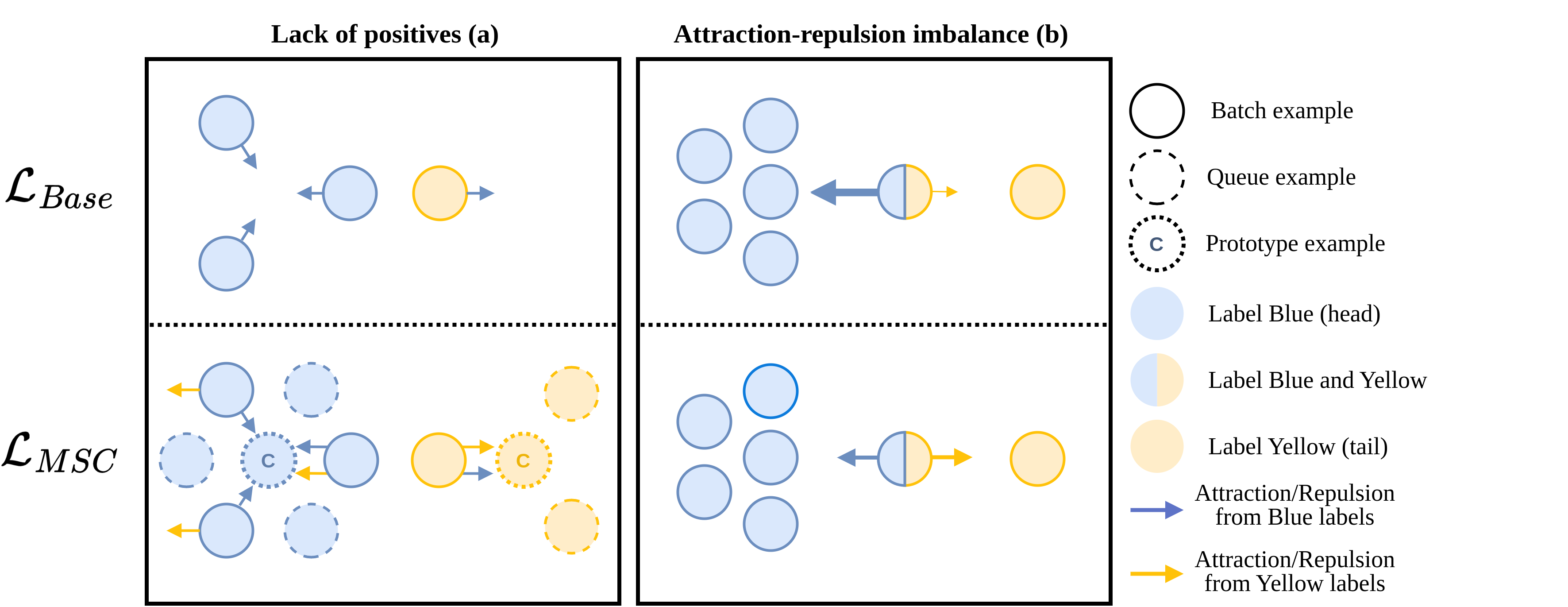}
\caption{Illustration of how ``lack of positives'' and ``attraction-repulsion imbalance'' problem are addressed by $\mathcal{L}_{Base}$  (classical contrastive loss for MLTC) and $\mathcal{L}_{MSC}$ (our proposed balanced Multi-label Supervised Contrastive loss). (a) Adding prototypes and a queue in $\mathcal{L}_{MSC}$ ensures a consistent positive pairing and expands positive and negative samples diversity. (b) Reweighting negative pairs addresses the imbalance between head and tail labels. For clarity, only the attraction/repulsion on the sample in the middle is depicted, without queue and prototypes. Color blue (respectively yellow) corresponds to a label in the head (respectively tail) of the distribution. }
 \label{fig:Issues}
\end{figure*}

\subsection{Contrastive Baseline $\mathcal{L}_{Base}$}
Before introducing our approach, we provide a description of our baseline for comparison, denoted as  $\mathcal{L}_{Base}$, and defined as follows:
\begin{equation*}
   \begin{split}
   &\mathcal{L}_{Base} = - \frac{1}{|B|} \sum_{\vz_i \in I_B} \frac{1}{N(i)} \\ 
   &\sum_{\vz_j \in I_B \backslash \vz_i} \frac{|\vy_i \cap \vy_j|}{|\vy_i \cup \vy_j|} \log\frac{\exp(\vz_i \cdot \vz_j /\tau)}{\sum_{\vz_k \in I_B \backslash \vz_i} \exp(\vz_i \cdot \vz_k /\tau)}
   \end{split}
   \label{eq:lbase}
\end{equation*}
This loss is a simple extension of the {\it SupCon} loss \cite{SupCon} with an additional term introduced to model the interaction between labels, corresponding to the Jaccard Similarity. $\tau$ represents the temperature, $\cdot$ represents the cosine similarity, and $N(i)$ is the normalization term defined as:
\begin{equation*}
   N(i) = \sum_{j \in B \backslash i} \frac{|\vy_i \cap \vy_j|}{|\vy_i \cup \vy_j|}
\end{equation*}
It is to be noted that $\mathcal{L}_{Base}$, does not consider the inherent long-tailed distribution of multi-label dataset, and that it is similar to other losses proposed in contrastive learning \cite{contrastive_MLTC, effectivemltcco}. We provide further details in Appendix \ref{appendix:base}. 
\subsection{Motivation}
Our work can be dissected into two improvements compared to the conventional contrastive loss proposed for MLTC. 

Each of these improvements aims to tackle the long-tailed distribution inherent in the data and alleviate concerns related to the absence of positive instances and the imbalance in the attraction-repulsion dynamics. These improvements are outlined as follows.


\paragraph{Lack of Positive Instances:}
We use a memory system by maintaining a queue $Q~=~\{\tilde{\vz}_j \}_{j \in {\{1, \ldots, K\}}}$, which stores the learned representations of the $K$ preceding instances from the previous batches obtained from a momentum encoder. This is in line with other approaches \cite{Moco,Moco-v2} that propose to increase the number of positive and negative pairs used in a contrastive loss. Additionally, we propose to incorporate a set of $L$ trainable label prototypes $C = \{\vc_i\mid i \in \{1,\ldots, L\}\}$. This strategy guarantees that each example in the batch has at least as many positive instances as the number of labels it possesses. 
These two techniques are particularly advantageous for the labels in the tail of the distribution, as they guarantee the presence of at least some positive examples in every batch.


\paragraph{Attraction-Repulsion Imbalance:} Previous work highlights the significance of assigning appropriate weights to the repulsion term within the contrastive loss \cite{zhu2022balanced}. 
In the context of multi-label scenarios, our proposal involves incorporating a weighting scheme into the repulsion term (denominator terms in the contrastive loss function), to decrease the impact of head labels. More details about attraction and repulsion terms introduced in \citet{dissecting} can be found in Appendix \ref{appendix:attraction}. For an anchor example $i$ with respect to any other instances $k\neq i$ in the batch and in the memory queue, we define the weighting of the repulsion term as:
\begin{equation}
g_i(\vz_k, \beta) =
\begin{cases}
    1 & \text{if } \vz_k \in C, \\
    \beta & \text{otherwise. }
\end{cases}    
\end{equation}

with $0<\beta<1$. This function assigns equal weights to all prototypes, allocating less weight to all other examples present in both the batch and the queue. 

In contrastive learning for mono-label multi-class classification, the attraction term is consistently balanced, as each instance is associated with only one class. While, in MLTC, a document can have multiple labels, some in the head and others in the tail of the class distribution. Our approach not only weights positive pairs based on label interactions but also considers the rarity of labels within the set of positive pairs. Instead of iterating through each instance, we iterate through each positive label of an anchor defining a positive pair, as an instance associated with this label. 

Figure \ref{fig:Issues} illustrates the influence of addressing the lack of positives and attraction-repulsion imbalance with our new multi-label contrastive loss, denoted as $\mathcal{L}_{MSC}$,  compared to the original supervised contrastive loss, $\mathcal L_{Base}$ on the exact same training examples in two different situations.

\subsection{Multi-label Supervised Contrastive Loss}
To introduce properly our loss function, we use the following notation: $H = I \cup Q$ represents the set of embeddings in the batch and in the queue; $\Delta(\vz_i) = \{ k\in [1,L] | y_i^k = 1\}$ represents the set of labels for example $i$; and $P(j, i) = \{\vz_l \in H| y_l^j = 1\} \backslash \vz_i$ represents the set of representations for examples belonging to label $j$, excluding the representation of example $i$.
Our balanced multi-label contrastive loss can then be defined as follows~:
\begin{equation}
    \mathcal{L}_{MSC} = \frac{1}{|B|} \sum_{i \in B} \ell(\vz_i) 
\end{equation}
where $\ell(\vz_i)$ is the individual loss for example $i$ defined as~:
\begin{equation}
    \begin{split}
        &\ell(\vz_i) = -\frac{1}{|\vy_i|} \sum_{j \in \Delta(\vz_i)}\frac{1}{N(i, j)} \sum_{\vz_l \in P(j,i)\cup \vc_j } \\
        &f(\vz_i, \vz_j) \log \frac{\exp(\vz_i \cdot \vz_l /\tau)}{\sum_{\vz_k\in H \cup C \backslash \vz_i} g_i(\vz_k, \beta)\exp(\vz_i \cdot \vz_k))}
    \end{split}
\end{equation}
$g_i(\vz_k, \beta$) are our tailored weights for repulsion terms defined previously. $f$ represents the weights between instances and $N(i, j)$ is a normalization term, both are defined as: 

\begin{equation}
f(\vz_i, \vz_j) =
\begin{cases}
    1 & \text{if } \vz_j \in C \\
    \frac{1}{|\vy_i \cup \vy_j|}& \text{otherwise.}
\end{cases}
\label{eq:f}    
\end{equation}

\begin{equation}
    N(i, j) =   \sum_{\vz_l \in P(j,i) \cup \vc_j}  f(\vz_i, \vz_l)
\end{equation}

This $f$ defined in equation \ref{eq:f} is build so that the equation coincides with the Jaccard similarity in scenarios where labels are balanced. 

It is to be noted that until now, the learning of a representation space for documents through a pure contrastive loss has remained uncharted. Despite numerous studies delving into multi-label contrastive learning, none have exclusively employed contrastive loss without the traditional BCE loss.

%% file: latex/4ExperimentationDetails.tex
\section{Experimental Setup}
\label{sec:Exps}
This section begins with an introduction to the datasets employed in our experiments.  Subsequently, we will provide a description of the baseline approaches against which we will compare our proposed balanced multi-label contrastive loss, along with the designated metrics.
\subsection{Datasets}
We consider the following three multi-label datasets.
\begin{enumerate}
    \item \textbf{RCV1-v2} \cite{rcv1-v2}: RCV1-v2 comprises categorized newswire stories provided by Reuters Ltd. Each newswire story may be assigned multiple topics, with an initial total of 103 topics. We have retained the original training/test split, albeit modifying the number of labels. Specifically, some labels do not appear in the training set, and we have opted to retain only those labels that occur at least 30 times in the training set. Additionally, we extract a portion of the training data for use as a validation set.
    \item \textbf{AAPD} \cite{sgm}: The Arxiv Academic Paper Dataset (AAPD) includes abstracts and associated subjects from 55,840 academic papers, where each paper may have multiple subjects. The goal is to predict the subjects assigned by arxiv.org. Due to considerable imbalance in the original train/val/test splits, we opted to expand the validation and test sets at the expense of the training set. 
    \item \textbf{UK-LEX} \cite{uklex_dataset}: United Kingdom (UK) legislation is readily accessible to the public through the United Kingdom's National Archives website\footnote{\url{ttps://www.legislation.gov.uk}}. The majority of these legal statutes have been systematically organized into distinct thematic categories such as health-care, finance, education, transportation, and planning.
\end{enumerate}
Table \ref{table:datasets} presents an overview of the main characteristics of these datasets, ordered based on the decreasing number of labels per example.
\input{tables/dataset}
\subsection{Comparison Baselines}
To facilitate comparison, our objective is to assess our approach against the current state-of-the-art from two angles. We first examine methods that focus on the learning of a robust representation, and then we assess approaches that are centered around BCE and its extensions.

\subsubsection{Baseline: Learning a good representation space}
\label{sec:421}
We assess our balanced multi-label contrastive learning by comparing it with the following loss functions that were introduced for learning improved representation spaces.

\noindent $\bullet~~ \mathcal{L}_{MLM}$, represents the classical masked language model loss associated with the pre-training task of transformer-based models \cite{roberta}. 

\smallskip

\noindent $\bullet~~ \mathcal{L}_{Base}$, serves as our baseline for contrastive learning, as presented in the previous section.
\smallskip

\noindent $\bullet~~ \mathcal{L}_{BQueue}$, corresponds to $\mathcal{L}_{Base}$ with additional positive instances using a queue.

\smallskip

\noindent $\bullet~~ \mathcal{L}_{BQProto}$, represents the strategy that involves integrating prototypes into the previous  $\mathcal{L}_{BQueue}$ loss function.

\medskip


\subsubsection{Standard loss function for Multi-Label}
\label{sec:422}
The second type of losses that we consider in our comparisons are based on BCE.

\noindent $\bullet~~ \mathcal{L}_{BCE}$, denotes the BCE loss, computed as follows~:
\begin{equation*}
    \begin{split}
   &\mathcal{L}_{BCE} = -\frac{1}{N} \sum_{i=1}^N \frac{1}{L} \sum_{j=1}^L \\
    &y_i^j\log( \hat{y}_i^j) + (1 - y_i^j)\log( 1 - \hat{y}_i^j)
    \label{eq:bce}
    \end{split}
\end{equation*}
where, $\{\hat{y}_i^1, ..., \hat{y}_i^L\}$ represent the model's output probabilities for the $i^{th}$ instance in the batch.
\smallskip

\noindent $\bullet~~ \mathcal{L}_{FCL}$, denotes the focal loss, as introduced by \citet{focal}, which is an extension of $\mathcal{L}_{BCE}$. It incorporates an additional hyperparameter $\gamma~\geqslant~0$, to regulate the ability of the loss function to emphasize over difficult examples.
\begin{equation*}
    \begin{split}
    &\mathcal{L}_{FCL} =- \frac{1}{N} \sum_{i=1}^N \frac{1}{L} \sum_{j=1}^L\\
    & y_i^j(1 - \hat{y}_i^j)^{\gamma}\log( \hat{y}_i^j)+(1 - y_i^j)( \hat{y}_i^j )^\gamma \log( 1 - \hat{y}_i^j)
    \end{split}
    \label{eq:focal}
\end{equation*}
\smallskip

\noindent $\bullet~~ \mathcal{L}_{ASY}$, represents the asymmetric loss function \cite{asymmetric} proposed to reduce the impact of easily predicted negative samples during the training process through dynamic adjustments, such as ’down-weights’ and ’hard-thresholds'. The computation of the asymmetric loss function is as follows:
\begin{equation*}
\begin{split}
    &\mathcal{L}_{ASY} = - \frac{1}{N} \sum_{i=1}^N \frac{1}{L} \sum_{j=1}^L \\ 
    &y_i^j(1 - s_i^j)^{\gamma^+}\log(s_i^j)(1 - y_i^j)+( s_i^j )^{\gamma^-}\log( 1 - s_i^j)
\end{split}
    \label{eq:asy}
\end{equation*}
with $s_i^j = \max(\hat{y}_i^j - m, 0)$.  The parameter $m$ corresponds to the hard-threshold, whereas $\gamma^+$ and $\gamma^-$ are the down-weights.
\input{tables/ablation}

\subsection{Implementation Details}
Our implementation is Pytorch-based\footnote{\url{https://pytorch.org}}, involving the truncation of documents to 300 tokens as input for a pre-trained model.
For AAPD, RCV1 datasets, we utilized the Roberta-base \cite{roberta} as the backbone, implementing it through Hugging Face's resources\footnote{\url{https://huggingface.co/roberta-base}}. For the UK-LEX dataset, we employed Legal-BERT, also provided by Hugging Face\footnote{\url{https://huggingface.co/nlpaueb/legal-bert-base-uncased}}. 
As common practice, we designated the [CLS] token as the final representation for the text, utilizing a fully connected layer as a decoder on this representation. Our approach involves a batch size of 32, and the learning rate for the backbone is chosen from the set $\{5e^{-5}, 2e^{-5}\}$. Throughout all experiments, we use AdamW  optimizer \citep{AdamW}, setting the weight decay set to $0.01$ and implementing a warm-up stage that comprises 5\% of the total training. For evaluating the representation space, we trained logistic regressions with AdamW separately for each individual label. To expedite training and conserve memory, we employed 16-bit automatic mixed precision. Additional details and the pseudocode of our approach are available in Appendices \ref{appendix:details} and \ref{appendix:algo} respectively.

The evaluation of results is conducted on the test set using traditional metrics in MLTC, namely the hamming loss, Micro-F1 score and Macro-F1 score \cite{enhancing_multi-label}.

%% file: tables/dataset.tex
\begin{table}
\centering
\setlength{\tabcolsep}{4pt}
\begin{tabular}{lcccccc}
\toprule[1.2pt] 
Dataset & $|\text{Train}|$ & $|\text{Val}|$ & $|\text{Test}|$ & ${L}$ & $\overline{L}$ & $\overline{W}$ \\  
\midrule
RCV1 & 19.7k & 3.5k & 781k & 91 & 3.2 & 241 \\
AAPD & 42.5k & 4.8k & 8.5k & 54 & 2.4 & 163 \\
UK-LEX & 20.0k & 8.0k & 8.5k & 69 & 1.7 & 1154 \\
\bottomrule[1.2pt] 
\end{tabular}
\caption{Datasets statistics. The table shows the number of examples (in thousands) within the training, validation, and test sets, as well as the number of class labels $L$, the average number of labels per example $\overline{L}$, and the average word count per document $\overline{W}$.}
\label{table:datasets}
\end{table}




%% file: tables/ablation.tex
\begin{table*}[t]
\centering
\begin{tabular}{lccc ccc}
\toprule[1.3pt] 
\multicolumn{1}{c}{} & \multicolumn{3}{c}{AAPD} & \multicolumn{3}{c}{RCV1} \\
\cmidrule(r){2-4} \cmidrule(lr){5-7}
\textbf{Loss} & $\mu$-F$_1$ & M-F$_1$& Ham & $\mu$-F$_1$ & M-F$_1$ & Ham \\
\cmidrule{1-7}
$\mathcal{L}_{MLM}$ & 63.86 & 45.62 & 28.48 & 80.06 & 58.42 & 13.5 \\
\cmidrule{1-7}
$\mathcal{L}_{Base}$ & 72.25 & 56.42 & 24.4 & 87.89 & 73.7 & 8.51 \\
$\mathcal{L}_{BQueue}$ & 72.73 & 57.92 & 24.15 & 87.56 & 72.9 & 8.72 \\
$\mathcal{L}_{BQProto}$ & 73.3 & 59.126 & \textbf{23.69} & 88.00 & 74.82 & 8.44 \\
$\mathcal{L}_{MSC}$ (ours) & \textbf{73.59} & \textbf{60.00} & 23.74 & \textbf{88.40} & \textbf{76.82} & \textbf{8.21} \\
\bottomrule[1.3pt] 
\end{tabular}
\caption{Evaluation of progressive complexity in contrastive loss functions. Micro-F1 ($\mu$-F$_1$), Macro-F1 (M-F$_1$), and Hamming Loss (multiplied by $10^3)$ metrics are averaged over nine values (three seeds and three temperatures  ${0.07,.1,.2}$) - except for $\mathcal{L}_{MLM}$ averaged on three seeds.}
\label{table:ablation}
\end{table*}

%% file: latex/5Results.tex
\section{Experimental Results}
\label{sec:Results}
We start our evaluation by conducting an ablation study, comparing various loss functions proposed for representation learning, as outlined in Section \ref{sec:421}.  Table \ref{table:ablation} summarizes these results obtained across various temperatures and seeds. The score achieved with $\mathcal{L}_{MLM}$ is merely 10 points lower in the Micro-F1 score compared to the best results, highlighting the effectiveness of the representation space found during the pre-training phase. Our approach primarily focuses on the Macro-F1 score, targeting the prevalent long-tailed distribution in MLTC data. As the table shows, each additional component we have introduced contributes around one point to the Macro-F1 score. Maintaining a balance between attraction and repulsion terms proves crucial, particularly for RCV1-v2, where it resulted in a 2-point improvement in the Macro-F1 score.
\input{tables/ft}

Our proposed loss function, $\mathcal{L}_{MSC}$, exhibited superior performance over the baseline $\mathcal{L}_{Base}$ for all metrics, emphasizing the importance of addressing both the 'Lack of Positive' issue and the 'Attraction-Repulsion Imbalance' for an optimal representation space. Throughout our experiments, setting the temperature to 0.1 consistently yielded the best results across all baselines. Consequently, we adopted this setting for all subsequent experiments.

\subsection{Comparison with standard MLTC losses}
Table \ref{table:all_datasets} presents a comparison of performance between the standard BCE-based loss functions outlined in Section \ref{sec:422} and our approach.
$\mathcal{L}_{MSC}$ outperforms all baselines in Macro-F1 score. The asymmetric loss function achieves comparable results only for the AAPD dataset, albeit with the worst score in other metrics.
Regarding Micro-F1, the performance of the $\mathcal{L}_{Base}$ is equivalent for the AAPD dataset and slightly better for RCV1-v2 and UK-LEX compared to the best score of the three standard losses. 
 These results suggest that supervised contrastive learning in MLTC can achieve comparable or even superior results compared to standard BCE based loss functions without the addition of another loss function.
\subsection{Fine-Tuning after Supervised Contrastive Learning}
To evaluate the quality of the representation space given by the contrastive learning phase, we explored the transferability of features through a fine-tuning stage. This study introduces two novel baselines: $\mathcal{L}_{Base-FT}$ and $\mathcal{L}_{MSC-FT}$, which are obtained by fine-tuning the representation learn with contrastive learning instead of doing a simple linear evaluation. In all cases, $\mathcal{L}_{MSC-FT}$ achieved superior results in both micro-F1  and macro-F1 scores compared to $\mathcal{L}_{Base-FT}$. These results show that the features learned with $\mathcal{L}_{MSC}$ are robust and offer an enhanced starting point for fine-tuning, in contrast to the traditional $\mathcal{L}_{MLM}$. Conversely, the performance of $\mathcal{L}_{Base-FT}$ was either worse or comparable to that of BCE, which underlies the benefits of our new loss function.

\subsection{Representation Analysis}
\label{sec:RepAnalysis}
To quantify the quality of the latent space learned by our approach, we evaluate how well the embeddings are separated in the latent space according to their labels using two established metrics : Silhouette score \cite{silhouette} and Davies–Bouldin index \cite{dbi}. These metrics collectively assess the separation between clusters and cohesion within clusters of the embeddings.

We treat each unique label combination in the dataset as a separate class to apply these metrics to the multi-label framework. Such expansion can potentially dilute the effectiveness of traditional clustering metrics by creating too many classes. To mitigate this, our analysis focuses on subsets of the most prevalent label combinations, retaining only half of the most represented label combination.  A detailed exploration of the impact of the size of the subset selection is provided in the Appendix \ref{appendix:RepAnalysis}.

Table \ref{tab:representation_analysis} presents our findings. A direct comparison between the baseline contrastive method $\mathcal{L}_{Base}$, and our proposed $\mathcal{L}_{MSC}$ method (prior to fine-tuning) reveals a significant enhancement in both metrics score. The integration of fine-tuning using BCE significantly enhances $\mathcal{L}_{Base}$ and $\mathcal{L}_{MSC}$ for both metrics, which demonstrates the effectiveness of the hybrid approach. Using our loss with fine-tuning is the only method able to surpass BCE in both metrics. This underscores its efficacy in creating well-differentiated and cohesive clusters in the latent space.
\input{tables/representation_analysis}

%% file: tables/ft.tex
\begin{table*}[t]
\centering
\begin{tabular}{lccc ccc ccc}
\toprule[1.5pt] 
\multicolumn{1}{l}{} & \multicolumn{3}{c}{AAPD} & \multicolumn{3}{c}{RCV1} & \multicolumn{3}{c}{UK-LEX} \\
\cmidrule(r){2-4} \cmidrule(lr){5-7} \cmidrule(l){8-10}
\cmidrule(r){2-4} \cmidrule(lr){5-7} \cmidrule(l){8-10}
\multicolumn{1}{l}{\textbf{Loss}} & $\mu$-F$_1$ & M-F$_1$ & Ham & $\mu$-F$_1$ & M-F$_1$ & Ham & $\mu$-F$_1$ & M-F$_1$ & Ham \\
\cmidrule{1-10}
\multicolumn{10}{c}{\textbf{Supervised Loss}} \\
\cmidrule{1-10}
$\mathcal{L}_{ASY}$ & 72.92 & 60.63 & 25.3 & 86.63 & 75.02 & 10.02 & 70.53 & 60.58 & 14.43 \\
$\mathcal{L}_{FCL}$ & 73.85 & 59.91 & 22.61 & 88.36 & 76.69 & 8.19 & 73.23 & 61.17 & \textbf{11.54} \\
$\mathcal{L}_{BCE}$ & 73.89 & 59.98 & \textbf{22.53} & 88.17 & 76.06 & 8.17 & 72.61 & 60.97 & 11.95 \\
\cmidrule{1-10}
\multicolumn{10}{c}{\textbf{Contrastive Loss}} \\
\cmidrule{1-10}
$\mathcal{L}_{Base}$ & 72.51 & 56.67 & 24.13 & 87.86 & 73.79 & 8.48 & 72.3 & 59.66 & 12.31 \\
$\mathcal{L}_{Base-FT}$ &73.09 & 58.55 & 23.61 & 88.41 & 76.08 & 8.18 & 72.45 & 60.66 & 12.23 \\
\cmidrule{1-10}
\multicolumn{10}{c}{\textbf{Ours}} \\
\cmidrule{1-10}
$\mathcal{L}_{MSC}$ & 73.84 & \textbf{60.75} & 23.72 & 88.54 & 77.05 & 8.12 & \textbf{73.5} & \textbf{62.06} & 11.83 \\
$\mathcal{L}_{MSC-FT}$ & \textbf{74.00} & 60.41 & 23.01 & \textbf{88.65} & \textbf{77.18} & \textbf{7.99} & 72.97 & 61.33 & 12.04 \\
\bottomrule[1.5pt] 
\end{tabular}
\caption{Comparative Analysis of multi-label loss functions. Metrics used are Micro-F1 ($\mu$-F$_1$), Macro-F1 (M-F$_1$), and Hamming Loss (multiplied by $10^3)$. $FT$ stands for fine-tuning.}
\label{table:all_datasets}
\end{table*}

%% file: tables/representation_analysis.tex
\begin{table}[t]
\centering
\begin{tabular}{lcc}
\toprule[1.2pt]
\textbf{Method} & \textbf{Sil $\uparrow$} & \textbf{DBI $\downarrow$} \\ 
\midrule
$\mathcal{L}_{MLM}$ & -0.14  & 2.83 \\ 
$\mathcal{L}_{BCE}$ & 0.15 & 2.02 \\ 
\midrule
$\mathcal{L}_{Base}$ & 0.07  & 2.23 \\ 
$\mathcal{L}_{Base-FT}$ & 0.13  & 2.00 \\ 
\midrule
$\mathcal{L}_{MSC}$ & 0.10  & 2.07 \\ 
$\mathcal{L}_{MSC-FT}$ & \textbf{0.16} & \textbf{1.98} \\ 
\bottomrule[1.2pt]
\end{tabular}
\caption{Clustering Metrics for different loss functions on $10^4$ embeddings from RCV1-v2 test set. Only 50\% of most represented label combinations are kept.}
\label{tab:representation_analysis}
\end{table}


%% file: latex/6Conclusion.tex
\section{Conclusion}
\label{sec:Conclusion}
In this paper, we have introduced a supervised contrastive learning loss for MLTC which outperforms standard BCE-based loss functions. Our method highlights the importance of considering the long-tailed distribution of data, addressing issues such as the 'lack of positives' and the 'attraction-repulsion imbalance'. We have designed a loss that takes these issue into consideration, outperforming existing standard and contrastive losses in both micro-F1 and macro-F1 across three standard multi-label datasets. Moreover, we also verify that these considerations are also essential for creating an effective representation space. Additionally, our findings demonstrate that initializing the model's learning with supervised contrastive pretraining yields better results than existing contrastive pre-training methods.

%% file: latex/7Limitation.tex
\section{Limitation}
Even though our approach demonstrates effectiveness in practice, it is subject to certain limitations, as outlined in this paper. \\
Firstly,  our approach inherits the typical drawbacks of contrastive learning, including a prolonged training phase relative to traditional methods and the necessity of a secondary step to evaluate the representation space with linear evaluation.
Secondly, our experiments were solely conducted using the base version of the pre-trained model, without exploring the behaviors of supervised contrastive learning in larger versions of these models. \\
Lastly, investigating data augmentation for long texts presents challenges due to their discrete nature. We did not explore data augmentation techniques, despite the fact that they are critical in contrastive learning. Further research in this area could yield insightful contributions for future work. \\

%% file: latex/8Appendix.tex
\section{Implementation details}
\label{appendix:details}
This section describes the implementation details of our framework in six parts: experimentation baselines, standard approaches, pretraining for contrastive learning, evaluating representation space, the fine-tuning stage and GPU budget.\\
\textbf{Common Process for All Experiments:} The dropout rate in the pre-trained model is set to $0.1$, and weight decay is excluded from bias and LayerNorm parameters. The learning rate for parameters, other than the backbone, is consistently set to $5e^{-5}$. Gradient Clipping is used with the parameter set to 1. No data augmentation is employed.\
\textbf{Specifics for standard approaches:} In the baseline, we employed the standard linear scheduler, and the number of epochs was selected from $\{10, 40, 80\}$. As is commonly practiced, we employed a linear scheduler. During training, the model with the best F1-micro score is kept for testing, while the model achieving the best average results (averaged over seeds) on validation is retained for testing part.
In the baseline, we tested the standard parameters. For the focal loss we set in all experiments $\gamma = 2 $ and for the Asymmetric loss we set $\gamma^+ = 0, \gamma^- =3 , m = 0.3 $. \\
\textbf{Contrastive Learning Pretraining} Contrastive Learning tends to converge to a better representation with more iterations, which is why we consistently set the number of epochs to 80 in all experiments. We assessed the representation space of three checkpoints and retained the best one for testing. The available checkpoints include the last checkpoint, the one with the lowest loss in training, and the one with the lowest loss in validation. The checkpoints with the best micro-F1 is kept. As a common practice for contrastive learning, a cosine scheduler is used. As in SupCon \citet{SupCon}, we use a projection head composed of two fully connected layers with ReLU as activation function:
$W_2 \cdot \text{ReLU}(W_1 \cdot x)$ where $W_1 \in \mathbb{R}^{h \times h}$ and $W_2 \in \mathbb{R}^{d \times h}$ where $h$ is the dimension of the hidden space and $d$ is set to $256$ in our experiments. As in SupCon the projection head is discarded to evaluate the representation space. For the hyperparameter, we set the size of the MoCo queue equal to $512$ and the momentum encoder is update with a momentum equal to 0.999 as in \citet{Moco}. Finally, in our experiments, we set $\beta$ to 0.1; this parameter was not subject to search.\\
\textbf{Details evaluating representation space:} To study the representation space, we employed AdamW \citet{AdamW} for training logistic regression on frozen model, without exploring alternative optimizers. For each label we trained logistics regression with learning rate in the set $\{1, 1e^{-1}, 1e^{-2}\}$ and weight decay in $\{1, 1e{-1}, 1e^{-2}, 1e^{-4}, 1e^{-6}\}$ for a number of 40 epochs. To eliminate sensitivity to initialization, we trained 3 logistic regressions per label, and the output was computed as the mean probability.  For each label individually, the best parameters for the micro-F1 are kept.\\
\textbf{Fine-Tuning details:} When the best model checkpoint obtained by supervised contrastive learning is found, we discard the projection head and train a linear layer using BCE. The settings are the same as  "Common process for all experiments" and we searched
a learning rate in $\{5e^{-5}, 2e^{-5}\}$ and a number of epochs in $\{5, 10\}$.
\textbf{GPU budget:} In this section we will discuss on the GPU budget.
To start, it is crucial to note the number of parameters in the model utilized. We exclusively used base models, implying that the parameter count stands at 110 millions. For all experiments on AAPD and RCV1-v2 we used NVIDIA RTX A6000, and we used NVIDIA Quatro RTX A6000 for UK-LEX.
For the AAPD dataset, training a single model using contrastive learning requires 25 hours, while a fine-tuning step of 10 epochs takes 1 hour and 30 minutes. If we assume uniform time requirements across all datasets, the estimation suggests that all experiments will collectively take approximately 5000 hours.
\newpage

\section{Pseudo-code}
\label{appendix:algo}
\input{tables/algorithm}

\section{Comparative Analysis with Our Baseline and Past SCL for MLTC}
\label{appendix:base}
In this section, we compare our $\mathcal{L}_{Base}$ equation (refer to Equation \ref{eq:lbase}) with the two previously used loss functions in MLTC. 
\textbf{The Jaccard Similarity Contrastive Loss (JSCL)}: The $\mathcal{JSCL}$ introduced in \citet{effectivemltcco} shows significant resemblance, or is nearly identical, to our baseline. The primary difference lies in the position of the weight obtained through Jaccard similarity; in our approach, it is placed outside the logarithm.  If kept inside,  the coefficient has no impact on training  ($\log(ax)$ and $\log(x)$ have the same derivative), making the loss similar to defining a positive pair as any example that shares at least one label without weighting.

\begin{equation}
    \begin{split}
    &\mathcal{L}_{JSCL} = - \frac{1}{|B|} \sum_{\vz_i \in I} - \frac{1}{|B|} \\
    &\sum_{\vz_j \in A(i)}  \log\frac{|y_i \cap y_j|}{|y_i \cup y_j|}\frac{\exp(\vz_i \cdot \vz_j /\tau)}{\sum_{k\in A(i)} \exp(\vz_i \cdot \vz_k /\tau)}
    \end{split}
    \label{eq:jscl}
\end{equation}

\textbf{Contrastive Learning Multi-label:} The other loss function for SCL in MLTC called $\mathcal{L}_{con}$ in \cite{contrastive_MLTC} aimed to enhance the representation specifically for the utilization of K-Nearest Neighbors (KNN) algorithms. The primary distinction from our baseline lies in the similarity measure, utilizing distance, motivated by the application of KNN. Additionally, rather than employing Jaccard similarity, the authors utilized the conventional dot product.
$\mathcal{L}_{con}$ can be written as follows:

\begin{equation}
    \begin{split}
    &\mathcal{L}_{Con} = - \frac{1}{|B|} \sum_{\vz_i \in I} \frac{1}{C(i)} \\ 
    &\sum_{\vz_j \in A(i)} \langle y_i, y_j \rangle \log\frac{\exp(-d(\vz_i, \vz_j) /\tau)}{\sum_{k\in A(i)} \exp(-d(\vz_i, \vz_k) /\tau)}
    \end{split}
    \label{eq:lbasecon}
\end{equation}
$C(i)$ represents a classical normalization term like $N(i)$ and $d$ is a distance function. We observe that our contrastive baseline, $\mathcal{L}_{Base}$, exhibits significant similarity, requiring only minor modifications, thereby establishing it as a fair baseline.

\section{Clustering Quality Across Diverse Multi-Label Embeddings Proportions}
\label{appendix:RepAnalysis}
\begin{figure*}[h]
 \centering
\includegraphics[scale=.4]{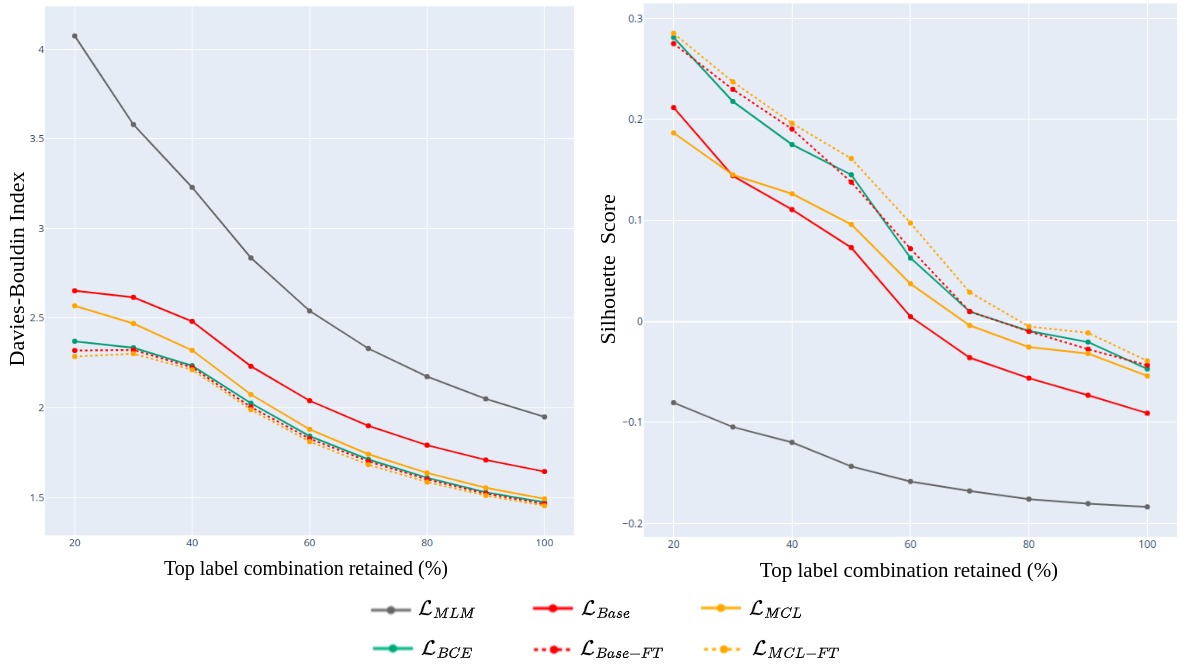}
\caption{Clustering quality metrics of different approaches across top classes retained.}
 \label{fig:clustering}
\end{figure*}

To apply clustering evaluation metrics such as the Silhouette score or the Davies-Bouldin index to multi-label embeddings, it is necessary to create one class for each unique multi-label combination, resulting in up to $2^L$ classes. Although $50$\% of these were retained in Table \ref{tab:representation_analysis}, we now explore a more general scenario by varying this proportion as reported in Figure \ref{fig:clustering}.

Our approach, $L_{MSC}$, consistently outperforms $L_{Base}$, except for a single proportion value of $20$\%, for Silhouette score. This could be attributed to the fact that our approach attempts to address the tail labels, which are typically discarded when keeping smaller proportions of top label combination.

\section{Attraction and Repulsion Term}
In this section, we define the classical \emph{SupCon} loss\cite{SupCon} as $\mathcal{L}_{SC}$. Given all instances representation Z of a batch with their corresponding class Y. The paper \citet{dissecting} shows that:
\label{appendix:attraction}
\begin{equation}
\begin{split}
     &\mathcal{L}_{SC}(Z;Y,B,y) \geq \sum_{i\in B_y} \log(|B_y| -1 + \\
     &|By^C|\exp(S_{att}^i(Z, Y, B, y) + S_{rep}^i(Z, Y, B, y)))
\end{split}
\end{equation}

Where:
\begin{equation}
 S_{att}^i(Z, Y, B, y) = - \frac{1}{|B_y| - 1}\sum_{j \in B_y \textbackslash i}
 \langle \textbf{z}_i, \textbf{z}_j \rangle
\end{equation}

\begin{equation}
\begin{split}
    S_{rep}^i(Z, Y, B, y) = \frac{1}{|B_y^C|} \sum_{j \in B_y^C} \langle \textbf{z}_i, \textbf{z}_j \rangle
\end{split}
\end{equation}
The set \( B_y^C \) denotes the indices of instances that do not possess the class \( y \), while \( B_y \) represents the indices of instances with the class \( y \).
\citet{zhu2022balanced} proposes the normalization of \(S_{rep}^i(Z, Y, B, y)\) involves re-weighting the denominator to achieve balance influence of classes. The attraction term \(S_{att}^i(Z, Y, B, y)\) relies on the numerator only, yet it can be adjusted by applying different weights before the logarithm.

\section{Study of $\mathcal{L}_{MulCon}$ representation space}
\label{appendix:mulocn}
In this section, we explain the claim that the loss function $\mathcal{L}_{MulCon}$ proposed in \citet{Mulcon} converges to a trivial solution without BCE.
In this work, the author inserts to the input a label representation called $\mL \in \mathbb{R}^{L\times d}$ where $d$ is the dimension of the hidden space. The output is composed of one representation per labels called $\mZ \in \mathbb{R}^{L\times d}$ and $\vz_i^k$ is the representation of the $k^{th}$ label for $i^{th}$ element inside a batch.
For one input, $\mX$ their model f can be summarized as:
\begin{equation}
    f(\mX, \mL) = \mZ
\end{equation}
We redefined $I = \{\vz_j^i | y_j^i = 1, j \in \{1, ..., N\}, i \in \{1,...,L\} \}$ the set of all labels representation
which appears inside a batch and the set of positive instance for the $i^{th}$ label of the $j^{th}$ instance
$P(i, j) = \{\vz_k^i | \vz_k^i \in I, y_k^i = y_j^i,  k \neq j\}$.
Under these notations $\mathcal{L}_{MulCon}$ can be defined as follows:
\begin{equation}
    \begin{split}
    & \mathcal{L}_{MulCon} = \frac{1}{|I|} \sum_{\vz_j^i \in I} \frac{1}{|P(i, j)|} \sum_{\vz_k^i \in P(i, j)}\\
    &\sum_{\vz_k^i \in P(i, j)} \log\frac{ \exp(\vz_j^i \cdot \vz_{k}^i /\tau)}{\sum_{\vz_f^t \in I \backslash {\vz_j^i}} \exp(\vz_j^i  \cdot \vz_f^t/\tau)}
    \end{split}
\end{equation}
For this demonstration, we position ourselves in the same configuration as \citet{dissecting}:
\begin{enumerate}
    \item $f$ is powerful enough to realize any geometric arrangement of the representations.
    \item Our dataset is balanced in terms of labels.
\end{enumerate}

Under these assumptions, the $\mathcal{L}_{MulCon}$ attains its minimum with $\{\vz_j^i = \zeta_i\}_{i=\{1, ..., L\}}$ where $\{\zeta_i\}_{i \in \{1,...,L\}}$ the vertices of an origin-centered regular $L-1$ simplex \citet{dissecting}. The previous expression shows that the representation of each label collapses, which implies that the output of the model is a constant C equals to $[\zeta_1,..., \zeta_L]$. The output does not depend on the input, which implies that the loss converges to a trivial solution without BCE.

%% file: tables/algorithm.tex
\begin{algorithm}
  \caption{Algorithms}
  \KwIn{ $g_{\theta} = f_{\theta} \circ h_{\theta}$ pre-trained model and its non-linear projection head}
  \hspace{12mm}$g_{\theta_k}$: momentum encoder of $g_{\theta}$ ;\\
  \hspace{11mm} $p_{\Tilde{\theta}}$: Linear Layer\\
  \hspace{11mm} m: momentum;\\
  \hspace{11mm} $\tau$: temperature;\\
  \hspace{11mm} Queue: Moco Queue;\\
  \hspace{12mm} C: Prototypes

  \For{x in loader}{
    \tcc{Forward Encoder and Momentum Encoder}
    $q = g_{\theta_q}(x)$;
    
    $k = g_{\theta_k}(x).detach()$;
    
    \tcc{Compute loss function}
    $\mathcal{L}$(q, Queue, C, $\tau$).backward();

    \tcc{Update Parameters}
    update($\theta$), update(C);
    
    \tcc{Update Parameters Moco Encoder}
    $\theta_k = m*\theta_k+(1-m)*\theta$

    \tcc{Update Queue}
    enqueue(Queue, k);
    dequeue(Queue)
 }
 
 \tcc{Discard the projection head h}
 $\Tilde{g}_{\Tilde{\theta}} = f_{\theta}.freeze() \circ p_{\Tilde{\theta}}$
 
 \tcc{Evaluate the representation space}
   \For{x, y in loader}{
    \tcc{Forward}
    $\hat{y} = \Tilde{g}_{\Tilde{\theta}}(x)$
    
    \tcc{Compute Standard BCE}
    BCE(y, $\hat{y}$).backward();

    \tcc{Update Parameters}
    update($\Tilde{\theta}$)
 }
 \KwOut{$\Tilde{g}_{\Tilde{\theta}}$}
\end{algorithm}